\documentclass{article}

\usepackage{acronym}
\usepackage[export]{adjustbox}
\usepackage{todonotes}
\usepackage{amssymb}
\usepackage{amsmath,bm}
\usepackage{physics}
\usepackage{subcaption}
\usepackage{orcidlink}
\usepackage[capitalise]{cleveref}
\title{Measuring the Data}
\author{Ido Cohen
\thanks{Ido Cohen \orcidlink{0000-0001-8142-9092} orcid 0000-0001-8142-9092
       {\tt\small ido.coh@gmail.com}}
}
\date{\today}

\begin{document}
\maketitle

\begin{abstract}
{\bf{Measuring the Data}} analytically finds the intrinsic manifold in big data. First, Optimal Transport generates the tangent space at each data point from which the intrinsic dimension is revealed. Then, the Koopman Dimensionality Reduction procedure derives a nonlinear transformation from the data to the intrinsic manifold. Measuring the data procedure is presented here, backed up with encouraging results.
\end{abstract}
\paragraph{Keywords} Optimal Transport $|$ Koopman Operator Theory $|$ Dimensionality Reduction



\section*{Significance Statement}Data interpolation using optimal transport with dimensionality reduction derived from the Koopman theory yields the \emph{Measuring the Data} procedure. A scattered data set becomes a net on the data manifold where the edges are parametric curves derived from optimal transport. A tangent space at each data point is isomorphic to the intrinsic manifold, that isomorphism can be found by Koopman theory. The main milestones of this work are presented and backed up with encouraging results. Neural Network understanding, data retrieval, data compression, and data denoising are only a few of the applications stemming from \emph{Measuring the Data}.

\section{Introduction}
"Weighing the World” is the name Cavendish gave to his seminal work on gravity, which cleared the fog surrounding this area since Isaac Newton formulated the gravity law. His work opened a new era in gravity research. Here, \emph{Measuring the Data} aims at calculating analytically the intrinsic dimension of sparse data. 
    
Dimensionality reduction processes mostly rely on either the high density of the data or linearity. However, oftentimes, neither is true. This work provides a deep theory to calculate the intrinsic dimension accurately in sparse data with nonlinear analysis tools. It leverages the connection of dynamics and data points via \ac{OT}. From data points, \ac{OT} extracts a vector field, Koopman Regularization procedure reveals a parsimonious representation of the dynamic, which is the nonlinear transformation connecting the data to the intrinsic coordinates. Thus, with Koopman theory and \ac{OT} we circumvent the sparsity and nonlinearity structure of the data.

\section{Main Steps of "Measuring the Data"}
The main steps of \emph{"Measuring the Data"} are as follows. 1) find parametric curves on the manifold between close data points. 2) extract a tangent bundle on the data manifold. 3) evaluate the \ac{ID} according to these bundles. 4) find the intrinsic coordinates of the data manifold. The following list elaborates on these steps.
\begin{figure}[phtb!]
    
    \includegraphics[trim=95 410 90 70, clip,width=.8\linewidth,valign = t]{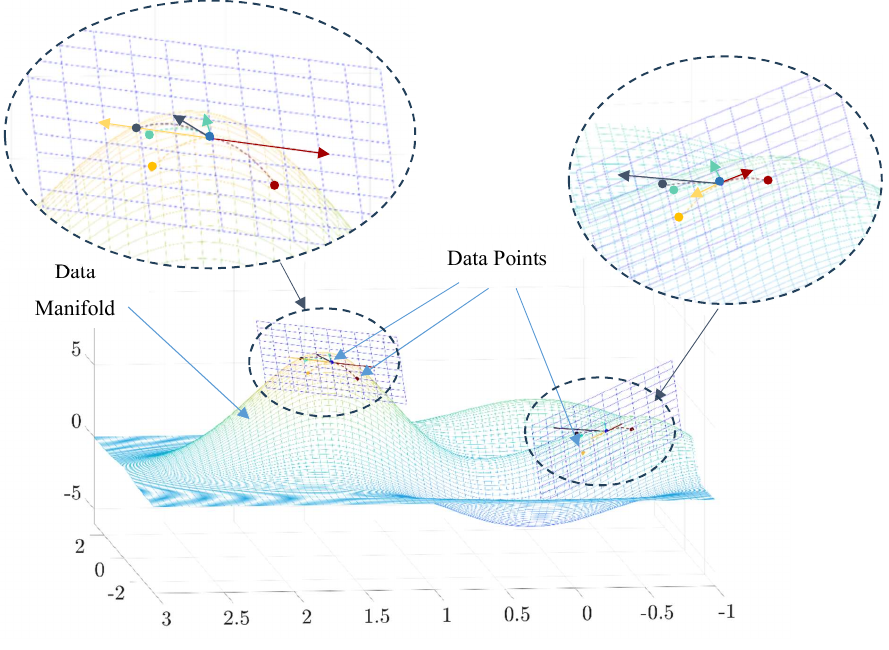}
    \caption{{\bf{Data Manifold}} – An illustration of “Measuring the Data”. At each point, we find a set of parametric curves representing data interpolation done by optimal transport. From these curves, find a tangent bundle at the data points. Then, extract the intrinsic dimension of the data manifold. Find the intrinsic coordinates. In the image, two neighborhoods (zoomed in) illustrate the concept of parametric curves, tangent bundles, and intrinsic dimension.}
    \label{Fig:figfig} 
\end{figure}


\begin{enumerate}
    \item {\bf{Parametric Curves on the Data Manifold}} \\ \ac{OT} is one of the ways to interpolate data by creating intermediate samples between two data points. If the intermediate points are valid data, they are laid on the intrinsic manifold. Thus, these interpolated data points form a parametric curve on the data manifold.
    
    \item {\bf{Tangent Bundle}} \\ Assume a data point is an intersection point of parametric curves (from Step 1) that connect it to its immediate neighbors (See the zoom in parts in \cref{Fig:figfig}). The velocity vectors of these curves tangent to the data manifold since the parametric curves are on this manifold. Thus, in the intersection points, one can find a tangent bundle to the data manifold. 
    
    \item {\bf{ID - Dimension of the Tangent Bundle Set}} \\ A tangent space to a data point on a $M$-manifold is isometric to $\mathbb{R}^M$. A tangent bundle is a subset of the tangent space; therefore, its dimension is the intrinsic dimension at most. Thus, the \ac{ID} can be extracted from the tangent bundles (generated in Step 2).
    
    \item {\bf{Intrinsic Coordinates}}\\ Given a tangent bundle (Step 2) and \ac{ID} (Step 3), one can find a linear mapping from the bundle to $R^M$. Assuming the data manifold is differentiable, there exists a (nonlinear) mapping from the data manifold to $\mathbb{R}^M$ where the Jacobian at each point is a linear mapping from the tangent bundle to $\mathbb{R}^M$. 
\end{enumerate}

\begin{figure}[phtb!]
    
    \includegraphics[trim=50 585 62 50, clip,width=.8\linewidth,valign = t]{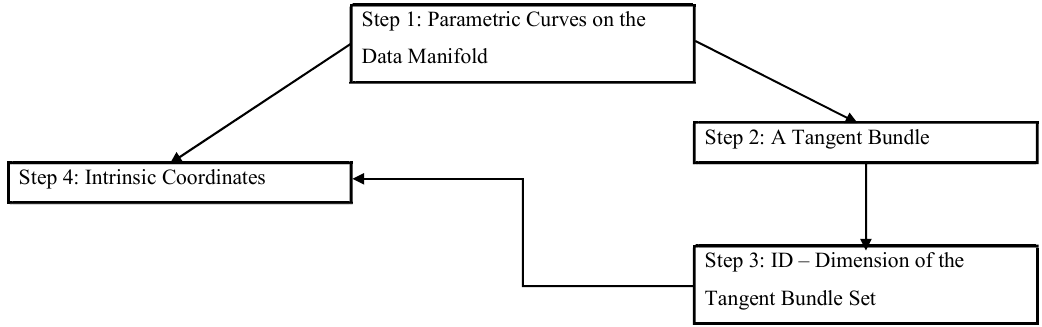}
    \caption{{\bf{“Measuring the Data” Flowchart}} –- Step 1: Create parametric curves from a data point to its immediate neighbors with optimal transport. Step 2: From these curves, find the tangent bundle at the intersection data points. Step 3: Find the intrinsic dimension from the tangent bundle. Step 4: Find the intrinsic coordinate from differential isomorphism from the data manifold to $\mathbb{R}^M$.}
    \label{Fig:chartFlow} 
\end{figure}

The process of “Measuring the Data” on 3d data points embedded in 2d manifold\\ is illustrated pictorially in \cref{Fig:figfig}. A general overview on Measuring the Data is given\\ in \cref{Fig:chartFlow}.

This work has two pillars. The first is the fact that the tangent space at a data point in $M$-manifold and $R^M$ are isomorphic (e.g. Remark 2.8.19 \cite{robbin2022introduction}). The second is the finite dimensionality of Koopman Eigenfunction Space \cite{cohen2024functional}. This discovery engendered an algorithm to find a parsimonious representation of dynamical systems from samples, termed \emph{Koopman Regularization} \cite{cohen2025koopmanregularization}. These two pillars intersect via optimal transport. Data can induce a vector field with \ac{OT} and Koopman Regularization provides a technique for dimensionality reduction. 



\section{Related Work}

Richard E. Bellman coined the term \emph{"curse of dimensionality"} in the context of dynamic programming \cite{bellman1961adaptive}. Today, though, it refers to the latent structure of the big data, which arises various challenges such as algorithms convergence and accuracy, storage, and when it comes to neural networks also in energy consumption and time training. Thus, dimensionality reduction in big data is of paramount importance, and finding the intrinsic dimension of a data set becomes crucial with time. The mathematical definitions of dimension and independent set are not univocal and have different meanings in different contexts, linear algebra, function spaces, topology, etc.. Concise yet meaningful data representation gets different forms depending on the application. However, an intuitive definition of the data dimension is the number of independent parameters describing the data \cite{fukunaga1971algorithm}. Namely, the data lies on a low dimension manifold where its coordinates are the independent parameters. 

Estimating the \ac{ID} in data mining stands at the core of many disciplines in a wide range of applications in data mining, classification, generative neural networks, characterization inference, and more. In general, algorithms estimate the \ac{ID} can be divided into three categories: local, global, and pointwise methods.  Local methods are based on the nearby geometry of a data sample to estimate the \ac{ID}. Global methods use the entire data for this estimation. Pointwise combines the local and the global for \ac{ID} estimating.  Reviewing these three methods can be exhaustive and since the \emph{measuring the data} procedure is a global method, the focus of this short survey is on the global methods. Global methods assume that the entire data lies on an $M$-manifold (where $M$ is constant) and roughly can be grouped into four categories, projection, multidimensional scale, fractal-based, and tangent bundle approximation. This survey is mostly based on the one in \cite{camastra2016intrinsic} but not only.

{\bf{The Projection Methods}}\\
This method searches for the most informative subspace to project the data on it with minimum information loss. The \ac{PCA} method and its variants project the data on the direction in which the variance is highest. Although easy to implement, \ac{PCA} overestimates the \ac{ID}, very sensitive to noise, and requires manual intervention to set a threshold to choose the main components and those to neglect. To overcome these drawbacks, a series of variants were suggested, Probabilistic \ac{PCA} \cite{tipping1999probabilistic}, Bayesian \ac{PCA} \cite{Bishop1998Bayesian}, Sparse PCA \cite{zou2006sparse}, and Sparse Probabilistic \ac{PCA} \cite{guan2009sparse}. These variants try to deal with the overestimated \ac{ID} problem or to set a robust strategy to decide what component to include. However, these main issues remained without proper answers. The nonlinear \ac{PCA} outperforms \cite{karhunen1994representation} the linear \ac{PCA}, however, the projections on curves and surfaces are not optimal \cite{malthouse1998limitations}.

{\bf{Multidimensional Scaling}}\\
These methods approximate a chart preserving geodesics \cite{bennett1969intrinsic,kruskal1964multidimensional,sammon2006nonlinear,Shepard1972Multidimensional}. These are exhaustive algorithms where the \ac{ID} is chosen from a repetitive gradient descent process to minimize the stress function and find the optimum achievable dimensional reduction. \ac{MDS} is easy and fast to implement if the data dimensionality is not high, which is very often not.

{\bf{Fractal Based Methods}}\\
Unlike the other ones, \ac{FBM} approximate non-integer \ac{ID}. K\'egl’s algorithm \cite{kegl2002intrinsic} and its extension \cite{raginsky2005estimation} leverage the equivalent of independent data points on a net and Kolmogorov capacity \cite{EdwardOtt2002CiDS} to estimate \ac{ID}. Despite its low time consumption, K\'egl’s algorithm is not robust to high-dimensional data. Grossberger-Procaccia algorithm \cite{grassberger1983measuring} and \cite{fan2009intrinsic,kalantana2013intrinsic} approximate the Correlation Dimension, the lower bound of Kolmogorov capacity. For example, Takens in \cite{Takens1985numerical} uses the maximum likelihood to approximate this lower bound, and an extension to the Grossberger-Procaccia algorithm copes with high \ac{ID} \cite{camastra2002estimating}.

{\bf{Tangent Bundle Approximation}}\\ 
\ac{TBA} is a growing group of algorithms that seek the diffeomorphism from the data space to the latent one \cite{chen2018sparse,silva2003geometric,silva2004motion,silva2005non,silva2005selecting}. Based on local \ac{PCA} or \ac{LASSO}, this method approximates the tangent space of the data manifold. Then, the next step is to find a chart from the point of the data and its approximated tangent bundle to a lower dimension space. This method yields impressive results; however, it inherits the drawbacks of \ac{PCA} and \ac{LASSO}, such as unrobustness to noise. In addition, this method works only when the data samples are dense enough, which is not likely very often.


The aforementioned methods try to approximate the \ac{ID} based on the assumption of either highly dense data or linearity of the mapping from the data space to the latent space. Both are naive and far from reality. The high density and linearity assumptions yield overestimated \ac{ID} and high sensitivity to noise.

This work is under the umbrella of the global method approach. Through optimal transport, measuring the data connects the data structure to dynamics. Then, using the knowledge of Koopman theory and Riemannian manifold and geometry nonlinear diffeomorphism is formulated from the data space to the intrinsic manifold.

\section{Preliminaries}
The Koopman Operator theory and Optimal Transport are the two pillars of this work. Definitions, terms, and notations necessary to understand this work are listed below. 
\subsection{Notations and Definitions}
Let us summarize some basic natation regarding linear algebra and calculus in \cref{tab:Notations}.
\begin{table}[htp!]
    \begin{tabular}{l l}
       $\nabla$ & -- Gradient operator \\
       $\bm{f}$ & -- Bold letter refers to a vector (of variables or functions)\\
       $\bm{1}$& -- a vector where each entry is $1$\\
       $J_{f}(\bm{\bm{x}})$ & -- Jacobian matrix\\
    $\norm{\cdot}$ & -- Standard Euclidean norm\\
    $\dot{}$ & -- Time derivative\\
    $^T$& -- (superscript $T$) Transpose
    \end{tabular}
    \caption{Notations}
    \label{tab:Notations}
\end{table}

\paragraph{Functionally Independent Set} Given a set of functions from $\mathbb{R}^N$ to $\mathbb{R}$, it is said to be a functionally independent set if the Jacobian is a full-rank matrix (e.g. defined in \cite{zorich2016mathematical} Section 8.6).

\subsection{Dynamics} 
\paragraph{Dynamical System} Consider an $N$ dimension dynamical system,
\begin{equation}\label{eq:dyn}
    \dot{\bm{x}}(t)=P(\bm{x}(t))
\end{equation}
where $\bm{x}\in\mathbb{R}^N$ and $P:\mathbb{R}^N\to \mathbb{R}^N$ and $P\in C^1(\mathbb{R}^N )$. 

\paragraph{Measurement} A measurement is a function from the phase space to $\mathbb{C}$. 

\paragraph{Unit Velocity Measurement}
A unit velocity measurement is a measurement, denoted by $m(\bm{x})$, admitting 
\begin{equation}\label{eq:UVM}
    \nabla^T m (\bm{x})P(\bm{x})=1.
\end{equation}

\paragraph{Koopman Operator} The operator of Koopman, $\mathcal{K}_{\tau}$, acts on a measurement as follows
\begin{equation}
    \mathcal{K}_\tau (g(x(t)))=g(x(t+\tau)).
\end{equation}
\paragraph{\acf{KEF}} A Koopman Eigenfunction, denoted by $\varphi$, associated with an eigenvalue $\mu$, is a measurement that admits the following relation
\begin{equation}
    \mathcal{K}_\tau (\varphi (x(t)))=\mu \varphi (x(t))    
\end{equation}
where $\mu\in\mathbb{C}$. As a consequence of the linearity and semi-group attributes of the Koopman operator, a \ac{KEF} admits the following \ac{ODE}
\begin{equation}\label{eq:KEF}
    \frac{d\varphi(t)}{dt}=\lambda \varphi (t),    
\end{equation}
where $\lambda=\dfrac{1}{\tau}\ln{\mu}$.

\paragraph{\acf{KPDE}} More general definition, related to Koopman Operator is the \ac{KPDE} formulated as
\begin{equation}\label{eq:KPDE}
    \nabla^T \phi(\bm{x})P(\bm{x})=\lambda\phi(\bm{x}).    
\end{equation}
This \ac{PDE} is the result of applying the chain rule on \cref{eq:KEF}. A solution of \cref{eq:KPDE} is a solution of \cref{eq:KEF} but not vice versa.

\paragraph{Conservation Law} A conservation law of a dynamic is a nontrivial solution to \cref{eq:KPDE} associated with $\lambda = 0$, denoted as $h(\bm{x})$,
\begin{equation}
    \nabla^T h(\bm{x})P(\bm{x})=0.    
\end{equation}
Note that, there are only $N-1$ functionally independent conservation laws \cite{cohen2024functional}.

\paragraph{General Form of \ac{KPDE}} The general form of the solution to \ac{KPDE}, \cref{eq:KPDE}, 
\begin{equation}
    \phi(\bm{x}) = f(h_1(\bm{x}),\ldots,h_{N-1}(\bm{x}))e^{\lambda m(\bm{x})}
\end{equation}
where $f$ is a differential function from a relevant subset in $\mathbb{R}^{N-1}$ to $\mathbb{C}$,  $\{h_i(\bm{x})\}_{i=1}^{N-1}$ is an functionally independent set of conservation laws, and $m(\bm{x})$ is a unit velocity measurement and $\lambda\in\mathbb{C}$ (discussed in \cite{cohen2024functional}). 

Note that, a Koopman eigenfunctions and a unit velocity measurement admit the following relation
\begin{equation}\label{eq:unitKoopmanR}
    m(\bm{x})=\frac{1}{\lambda} \log{\phi(\bm{x})}.
\end{equation}
\paragraph{Minimal Set of \acp{KEF}} A set of \acp{KEF} $\{\phi_i\}_{i=1}^K$ is a minimal set if this set is functionally independent with maximal cardinality. If the dynamic representation is not redundant, the cardinality is $N$.

\paragraph{Minimal set of Unit Velocity Measurement} The minimal set of \acp{KEF} extends the foliation theory in topology \cite{lawson1974foliations} and the Frobenius theory \cite{tall2011flow}. Given a minimal set of Koopman eigenfunctions, one can extract a minimal set of unit velocity measurements by applying the relation \cref{eq:unitKoopmanR}. Thus, naturally, a functionally independent set of \acp{KEF} leads to a minimal set of unit velocity measurement forming a set of functionally independent flow-box coordinates.

\subsection{Koopman Regularization}
\ac{KR} is a constrained optimization-based algorithm to discover the governing equations from samples \cite{cohen2025koopmanregularization}. \ac{KR} finds a minimal set of unit velocity measurements and thus formulates a parsimonious representation of the dynamics at hand. This set yields the following system of partial differential equations 
\begin{equation}\label{eq:unitPDES}
    J_m(\bm{x})P(\bm{x})=\bm{1}.
\end{equation}
If the cardinality is maximal then $J_m{(\bm{x})}$ is a full-rank $N\times N$ matrix, and the governing equations can be discovered by
\begin{equation}\label{eq:UVMrecon}
    P(\bm{x})=J_m(\bm{x})^{-1}\bm{1}.
\end{equation}

This equation reveals the dynamics when the cardinality if $N$. However, if the samples are taken from a redundant representation, $J_m{(\bm{x})}$ is not a square matrix, i.e. the cardinality is $K(<N)$, and \cref{eq:UVMrecon} is not valid. On the other hand, the relation in \cref{eq:unitPDES} is valid for any cardinality, meaning that at any point the vector filed is a linear combination of $\{\nabla m_i(\bm{x})\}_{i=1}^K$ . In this case, the dynamical system reconstruction gets the form of
\begin{equation}\label{eq:DRVFRE}
    P(\bm{x}) = \sum_{i=1}^K\alpha_i(\bm{x})\nabla m_i(\bm{x})
\end{equation}
where $\{\alpha_i\}_{i=1}^K$ are real function acting on the state-space and $\{m_i\}_{i=1}^K$ is a functionally independent set of unit velocity measurements. 
\subsubsection{Loss Function}
The formulation of partial differential equations as an optimization problem depends on the cardinality. Let us consider the following functional
\begin{equation}\label{eq:funcLoss1}
\begin{split}
    \mathcal{L}(\bm{m})=\norm{J_m(\bm{x})P(\bm{x})-\bm{1}}^2
\end{split}.
\end{equation}
A set of measurements zeroing this functional under the constrain that $\{m_i\}_{i=1}^N$ is functionally independent set can reveal the vector field by applying \cref{eq:UVMrecon}.

If the cardinality is $K(<N)$, the vector field can be restored from \cref{eq:DRVFRE}. Thus, the functional to be minimized should contain a minimal set $\bm{m}(\bm{x})$ and the corresponding coordinates $\bm{\alpha}(\bm{x})$. A zeroing point of the following functional admits these conditions and restores the vector field
\begin{equation}\label{eq:funcLoss2}
\begin{split}
    &\mathcal{L}(\bm{\alpha},\bm{m})=\norm{P(\bm{x}) - \sum_{i=1}^K\alpha_i(\bm{x})\nabla m_i(\bm{x})}^2\\
    &\quad s.t. \,\, \{m_i\}_{i=1}^K\, \textrm{is functionally independent set}
\end{split}.
\end{equation}

\subsubsection{Optimization Method}  Optimization problems in variational calculus in general and constrained problems in particular are well studied \cite{OPTbao2004computing, OPTbao2003ground, OPTcaliari2009minimisation, OPTCOHEN20181138, OPTdemyanov2011exact, OPTdem2004exact, OPTekeland1999convex, OPTgarcia2001optimizing}. To solve the problems in \cref{eq:funcLoss1,eq:funcLoss2} the author used the barrier method \cite{OPTnesterov2018lectures}. In this method, the initial condition is in the feasible region (admitting the constraints) and the gradient descent flow is enforced to stay in this region. For more detail the reader is reffered to \cite{cohen2025koopmanregularization}

\subsection{Optimal Transport}
Monge formulated the optimal transport problem as reorganizing pile of mass with minimum effort \cite{monge1781memoire}. Since then this problem gets different applications such as merchandise distribution, computer graphics, and data interpolation\cite{diaz2024data,feng2024improving}. In the context of data mining, one can formulate \ac{OT} as follows. 

Let $f_1(x)$ and $f_2(x)$ two data sample from the data set $\mathcal{D}$. The data points $f_1, f_2$ can be images, signals or any quantity vary with some index $x$ (for example time or space). The aim is to find a mapping $T$ reorganizing the index $x$ such that 
\begin{equation}\label{eq:OTcond}
    f_1(x)=f_2(T(x))\abs{T'(x)}
\end{equation}
at any value of $x$ where $T'(x)=\frac{dT}{dx}$ and brings to infimum the following cost function
\begin{equation}\label{eq:OT}
    \inf_{T:X\to X} \int_X c(x,T(x))dx
\end{equation}
where $c(x,y)$ is the cost to pass an item from $x\in X$ to $y\in X$. 

Let us denote the solution of \cref{eq:OT} under the condition of \cref{eq:OTcond} as $T_{1,2}(x)$. Thus, the mapping $T_{1,2}(x)$ solving this optimization problem induces a smooth transmission from $f_2$ to $f_1$ and also defines intermediate points between these points. An intermediate point can be defined as
\begin{equation}\label{eq:parametricCurve}
    f_{1,2}(x,t)=f_2 (x(1-t)+t T_{1,2}(x))\abs{(1-t)+tT_{1,2}'(x)}
\end{equation}
where $0\leq t\leq 1$. When $t=0$ $f_{1,2}(x,0)=f_2(x)$ and $t=1$ $f_{1,2}(x,1)=f_1(x)$. For more theoretical background, see \cite{thorpe2018introduction,levy2018notions}. The following section elaborates on the building blocks in \cref{Fig:chartFlow}.

\section{Method -- Measuring the Data}
\subsection*{Step 1: Parametric Curves on the
Data Manifold -- From Data Points to Dynamical Systems}
Given two data samples, optimal transport can generate intermediate data points that are assumable on the data’s manifold (see e.g. \cite{thorpe2018introduction}). The intermediate points are on a parametric curve in the data manifold (\cref{eq:parametricCurve}). The derivation of the curve with respect to the parameter yields tangents of the manifold along the curve. Thus, at the point $f_2$, one can derive a tangent on the parametric curve connecting to $f_1$ as follows 
\begin{equation}\label{eq:TangAt2}
    V_{1,2}=\frac{d}{dt} f_{1,2}(t)\eval_{t=0}=f'_2 (x)(T_{1,2}(x)-x)+f_2 (x)(-1+T_{1,2}' (x)),
\end{equation}
where $T_{1,2} (x)$ is the solution of \cref{eq:OT} admitting the condition formulated in \cref{eq:OTcond}. The vector $V_{1,2}$ is the velocity of the parametric curve connecting $f_2(x)$ to $f_1(x)$ at the point $f_2(x)$. Under the assumption the curve in on the data manifold, meaning, each intermediate point is valid, the velocity $V_{1,2}$ is tangent to the data manifold.

\subsection*{Step 2: A Tangent Bundle -- Finding the Tangent Bundle}
Now, let us focus on a specific data point, $f_0(x)$, and its immediate neighbors $\{f_i(x)\}_{i=1}^k$. With optimal transport, one can generate a set of $k$ tangents at the point $f_0(x)$ when the $i$th tangent is associated with the curve connecting $f_0(x)$ to $f_i(x)$. If the ID is less than $k$, then this tangent set is linearly dependent. Whilst the category of \ac{TBA} only approximates the tangent bundle space, this work analytically finds the tangent space. From \cref{eq:TangAt2} in Step 1, the tangent bundle at $f_0$ can be calculated as
\begin{equation}\label{eq:TB}
    V_{i,0}=\frac{d}{dt} f_{i,0}(t)\eval_{t=0}=f'_0 (x)(T_{i,0} (x)-x)+f_0 (x)(-1+T_{i,0}' (x)).
\end{equation}
This bundle is part of the tangent space at the point in question. Thus, its dimension is at most the dimension of the tangent space which equals the intrinsic manifold dimension.

\subsection*{Step 3: ID – Dimension of the
Tangent Bundle Set}
The local dimension of the data manifold at the point $f_0(x)$ is the dimension of the tangent space. Thus, the local 
dimensionality at $f_0(x)$ can be calculated by applying \ac{SVD} to the associated tangents set, \cref{eq:TB} in Step 2. This technique is not new and does not have a specific benefit, however, since the tangents were got from optimal transport it is assumed to be more accurate. By the end of this step, the \ac{ID} is calculated from the exact local tangent sets.

\subsection*{Step 4: Intrinsic Coordinates -- Finding Intrinsic Manifold Coordinates with Koopman Eigenfunction Space}
Now, the last step is to find the diffeomorphism from the data in $\mathbb{R}^N$ to the intrinsic manifold $\mathbb{R}^M$ which define the coordinate system of the intrinsic manifold. Given the ID is $M$, the diffeomorphism is from the form or
\begin{equation}\label{eq:intrinsicCoordinates}
    \bm{\phi}(\bm{f})=\begin{bmatrix}
        \phi_1(\bm{f})\\
        \vdots\\
        \phi_M(\bm{f})
    \end{bmatrix}.
\end{equation}
where $\bm{f}\in\mathbb{R}^N$. Based on the identity mentioned in Remark 2.8.19 \cite{robbin2022introduction}, there exists a diffeomorphism from the tangent space in the data to the intrinsic manifold. Hence, from the intrinsic coordinates (\cref{eq:intrinsicCoordinates}) one can derive the mapping between the tangent spaces.

Let $V_{1,0}\in\mathbb{R}^n$ be the tangent on curve from $\bm{f}_0$ to $\bm{f}_1$ at the point $\bm{f}_0$, then the linear mapping from this vector to the tangent space in the intrinsic manifold is
\begin{equation}
    d\bm{\phi}(\bm{x})V_{1,0}=v_{1,0}
\end{equation}
where $v_{1,0}\in\mathbb{R}^M$ is the vector field in the intrinsic manifold. The differential formulation of this expression is
\begin{equation}\label{eq:diffeo}
    J_{\bm{\phi}}(\bm{f})\eval_{\bm{f}=\bm{f}_0}V_{1,0}=\begin{bmatrix}
        \nabla^T\phi_1(\bm{f})\\
        \vdots\\
        \nabla^T\phi_M(\bm{f})
    \end{bmatrix}V_{1,0}=\bm{v}_{1,0}
\end{equation}
where $J_{\bm{\phi}}(\bm{f})\in \mathbb{R}^{M\times N}$. This mapping is invertible. A tangent in the data space belongs to the row space of $J_{\bm{\Phi}}$, thus, the vector in $\mathbb{R}^N$ tangent to the data manifold is a linear combination of the coordinate's gradient, more formally,
\begin{equation}\label{eq:vectDR}
    V_{1,0}=\sum_{i=1}^M \alpha_{1,0}^{(i)}(\bm{f})\eval{\nabla \phi_i(\bm{f})}_{\bm{f}=\bm{f}_0}
\end{equation}
where $\alpha_{1,0}^{(i)}$ is the $i$th component (a smooth real function). 

The coordinate system $\{\phi_i\}_{i=1}^M$ is a chart of the neighborhood of $\bm{f}_0$. On the other hand, the expression in \cref{eq:vectDR} coincides with the dimensionality reduction of dynamical system as shown in Koopman Regularization section ( \cref{eq:DRVFRE}). 

\section{Results}
\begin{figure}[phtb!]
    \centering 
    \includegraphics[trim=0 0 0 0, clip,width=.9\linewidth,valign = t]{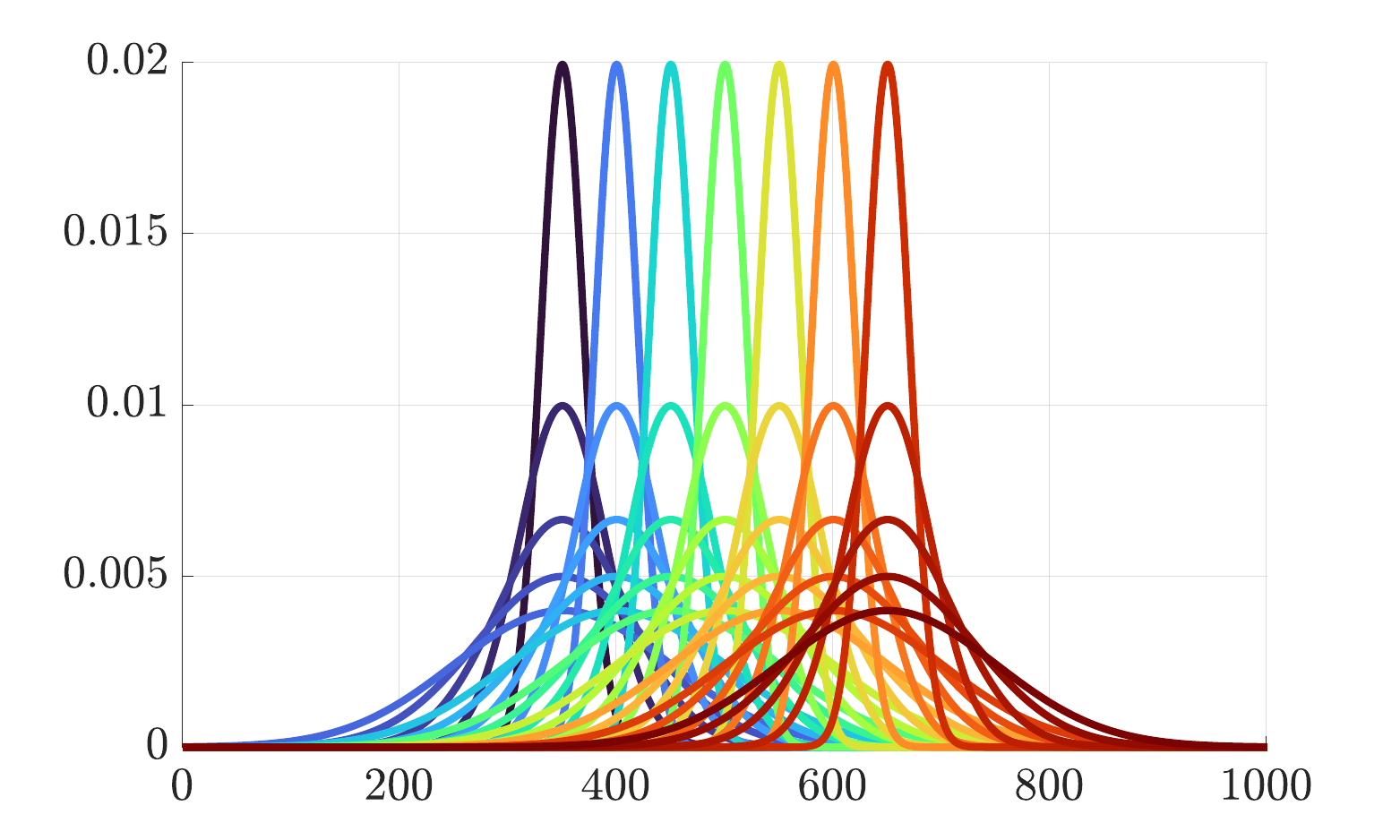}
    \caption{{\bf{Data Set}} -- A set of $35$ sampled Gaussian functions with different means and variance. Each function is sampled uniformly $1001$ times.}
    \label{Fig:toyExampleDataSet} 
\end{figure}

In what follows, a simple example emphasizes and demonstrates the merits of the \emph{Measuring the Data} procedure when it is applied to sparse and nonlinear data sets. Let the data set be one dimensional Gaussian functions $\{f_i\}_{i=1}^k$ of the form of 
\begin{equation}
    f_i (x)=\frac{1}{\sqrt{2\pi\sigma_i}} e^{-\frac{(x-\mu_i )^2}{2\sigma_i^2}}
\end{equation}
with $\mu_i$ and $\sigma_i$ as parameters. Then, these functions are equispaced sampled at the points $x=0,\,1,\,\ldots,\,1000$. The set is pictorially depicted in \cref{Fig:toyExampleDataSet}. There are $35$ data points (in $\mathbb{R}^{1001}$) when $\mu_i=350,\,400,\,\ldots,\,650$ and $\sigma_i=20,\,40,\,\ldots,100$.  As a result, there are $35$ data points where each of them is a vector in $\mathbb{R}^{1001}$. It is clear, that the data set is on $2d$ manifold since there are two parameters describing it accurately, $\mu_i$ and $\sigma_i$.

Applying \ac{TBA} to this set probably does not lead to an accurate approximation of $ID$. The sparsity of this set does not allow one to approximate the tangents simply by subtracting between data points. Optimal transport creates a smooth transition between data points, which each intermediate data point is a Gaussian function. Hence, the vector in \cref{eq:TB} tangents to the data manifold. The Python implementation of optimal transport is used to find the tangent bundle at each data point \cite{flamary2021pot}. 

\begin{figure}[phtb!]
    \centering 
    \includegraphics[trim=50 300 50 275, clip,width=.95\linewidth,valign = t]{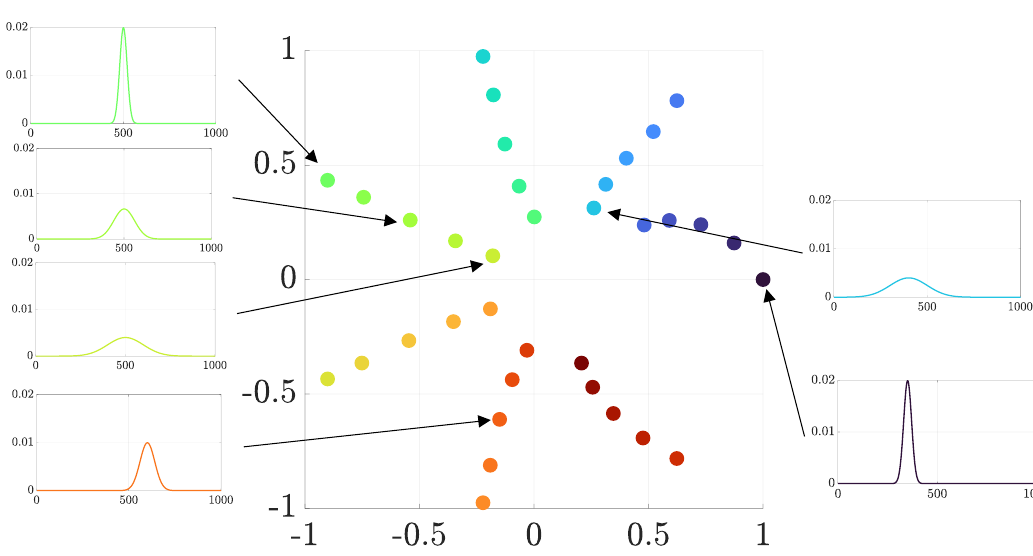}
    \caption{{\bf{Dimensionality Reduction with Measuring the Data}} Dimensionality Reduction with Koopman Regularization and Measuring the Data. Each Gaussian function (a vector in $\mathbb{R}^{1001}$) is mapped to two-dimensional space without information loss. The mapping is denoted with the corresponding color. The mean moves right in the data set and correspondingly the mapped points move anti-clockwise, and the distribution is relative to the distance from the origin.}
    \label{Fig:toyExampleResults} 
\end{figure}

After having the tangent bundles at each point, the local intrinsic dimension can be easily extracted with \ac{SVD}. In our case, the dimension is two and it is clearly found from the order of magnitude of the eigenvalues. Then a new coordinate system was trained according to \cref{eq:diffeo} by invoking the Koopman Regularization. The results are depicted in \cref{Fig:toyExampleResults}.

\section{Conclusion and Future Work}
The significance of this work cannot be ignored. We are surrounded by big data analysis and automatic machinery. Data processing and dynamics are everywhere. Each cell phone, each vehicle, and each medical imaging system must be equipped with strong and meaningful feature extraction, namely, dimensionality reduction. The data is gathered easily with time, and the tasks are more and more complicated, surveillance systems in airports, robotic systems in operations rooms, applications in control systems of unmanned aerial vehicles, and retrieval data in intelligence monitoring missions. In all these applications, either high dimensional dynamics or big data are involved. These two fields merge through mathematical discoveries from the last years. This work combines old and new methods, Koopman Operator (1932) and Optimal Transpose (1781), with cutting-edge research from Optimal Control as (\cite{flamary2021pot,martin2024data}) and the finite dimensionality of Koopman eigenfunction space (\cite{cohen2025koopmanregularization}). Here, optional applications and future works are discussed.

\paragraph{Data Compression} –- In the toy example, we saw sparse data from $\mathbb{R}^{1001}$ represented as two-dimensional coordinates. This result is expected as Gaussian distribution has two parameters, mean, and variance. This simple toy example demonstrates the ability of this algorithm to extract the local dimensionality of the data. Data from medical imaging is usually sparse with inherent resemblance. It is expected that due to \ac{OT}, one can measure the data and with this method, the belief is that the intrinsic coordinates of the data manifold represent meaningful parameters in the patient’s diagnosis. I assume compression to low dimensional representation may reduce some information (mostly noise) not necessarily lossless.

\paragraph{Data Denoising} –- Given the intrinsic coordinates and the original data, one can train a net that generates the data and intermediate samples as well. In this way, the meaning of these coordinates is clarified. The intrinsic coordinates of data may represent meaningful parameters, as presented in the toy example. On the other hand, fluctuations in the data and additive noise also have representations. Thus, this algorithm has the potential of restoration, not only in the “nonlocal” manner (for example, in image processing) but also in the meaning of different data points. 

\paragraph{Data Retrieval} –- The strength of this algorithm is in sparse data. Therefore, there is an expectation for outperforming in retrieval data (for example, see \cite{Weiss2008Spectral}). In addition, the suggested algorithm is a nonlinear mapping of the data into the intrinsic manifold. Thus, the intrinsic coordinates are expected to indicate certain features, and therefore, the retrieval yields more reasonable results. 

\paragraph{Neural Network Understanding} –- One of the great potentials of \emph{Measuring the Data} is to lift the curse of dimensionality from the data and from the neural network itself. A neural network is a highly dimensionally redundant dynamical system, as shown in \cite{brokman2024enhancing}. On the other hand, the data is also dimensionally redundant. Thus, this work paves the way for a new kind of optimization where the trained data and the neural network are processed as one unit to reduce redundant dimensionality.

\section*{Acknowledgment}
The author would like to thank Prof. Guy Gilboa, Prof. Gershon Wolansky, Dr. Eli Appleboim, and Dr. Yosef Meir Levi.


\section*{Acronym List}
\begin{acronym}
\acro{ID}[ID]{\emph{Intrinsic Dimension}}
\acro{PCA}[PCA]{\emph{Principal Component Analysis}}
\acro{MDS}[MDS]{\emph{Multidimensional Scaling}}
\acro{FBM}[FBM]{\emph{Fractal Based Methods}}
\acro{TBA}[TBA]{\emph{Tangent Bundle Approximation}}
\acro{OT}[OT] {\emph{Optimal Transport}}
\acro{KEF}[KEF]{\emph{Koopman Eigenfunction}}
\acro{KR}[KR]{\emph{Koopman Regularization}}
\acro{PDE}[PDE]{\emph{Partial Differential Equation}}
\acro{ODE}[ODE]{\emph{Ordinary Differential Equation}}
\acro{KPDE}[KPDE]{\emph{Koopman Partial Differential Equation}}
\acro{SVD}[SVD]{\emph{Singular value decomposition}}
\acro{LASSO}[LASSO]{\emph{Least Absolute Shrinkage and Selection Operator}}

\end{acronym}

\bibliographystyle{amsplain}
\bibliography{smartPeople}

\end{document}